\documentclass[letterpaper]{article} 
\usepackage{aaai24}  
\usepackage{times}  
\usepackage{helvet}  
\usepackage{courier}  
\usepackage[hyphens]{url}  
\usepackage{graphicx} 
\usepackage{natbib}  
\usepackage{caption} 
\usepackage{algorithm}
\usepackage{algorithmic}

\usepackage{newfloat}
\usepackage{listings}
\DeclareCaptionStyle{ruled}{labelfont=normalfont,labelsep=colon,strut=off} 
\floatstyle{ruled}
\newfloat{listing}{tb}{lst}{}
\floatname{listing}{Listing}
\usepackage{xspace}
\newcommand{\model}{\texttt{KG-TREAT}\xspace}
\usepackage{lipsum,booktabs}
\usepackage{adjustbox}
\usepackage{multicol}
\usepackage{multirow}
\usepackage{enumitem}
\usepackage{bm}
\newtheorem{assumption}{Assumption}
\newcommand\independent{\protect\mathpalette{\protect\independenT}{\perp}}
\def\independenT#1#2{\mathrel{\rlap{$#1#2$}\mkern2mu{#1#2}}}
\usepackage{amsmath,amsfonts,amssymb}

\title{KG-TREAT: Pre-training for Treatment Effect Estimation by \\Synergizing Patient Data with  Knowledge Graphs}
\author{
    Ruoqi Liu\textsuperscript{\rm 1},
    Lingfei Wu\textsuperscript{\rm 2},
    Ping Zhang\textsuperscript{\rm 1}
}
\affiliations{
    \textsuperscript{\rm 1}The Ohio State University\\
    \textsuperscript{\rm 2}Anytime.AI\\

    liu.7324@osu.edu, lwu@anytime-ai.com, zhang.10631@osu.edu
}

\usepackage{bibentry}

\begin{document}

\maketitle

\begin{abstract}
     Treatment effect estimation (TEE) is the task of determining the impact of various treatments on patient outcomes. Current TEE methods fall short due to reliance on limited labeled data and challenges posed by sparse and high-dimensional observational patient data. To address the challenges, we introduce a novel pre-training and fine-tuning framework, \model, which synergizes large-scale observational patient data with biomedical knowledge graphs (KGs) to enhance TEE. Unlike previous approaches, \model constructs \textit{dual-focus} KGs and integrates a deep bi-level attention synergy method for in-depth information fusion, enabling distinct encoding of treatment-covariate and outcome-covariate relationships. \model also incorporates two pre-training tasks to ensure a thorough grounding and contextualization of patient data and KGs. Evaluation on four downstream TEE tasks shows \model's superiority over existing methods, with an average improvement of 7\% in Area under the ROC Curve (AUC) and 9\% in Influence Function-based Precision of Estimating Heterogeneous Effects (IF-PEHE). The effectiveness of our estimated treatment effects is further affirmed by alignment with established randomized clinical trial findings.
\end{abstract}
\section{Introduction}
Treatment effect estimation (TEE), which identifies the causal effects of treatment options on patient outcomes given observational covariates, is a pivotal task in healthcare \citep{glass2013causal}. Yet, existing TEE methods \citep{shalit2017estimating, shi2019adapting, zhang2022can} are limited in both generalizability and accuracy due to their dependence on small, task-specific datasets that might not fully encompass the complex relationships among covariates, treatments, and outcomes.

\begin{figure*}[!t]
\centering
\includegraphics[width=0.99\textwidth]{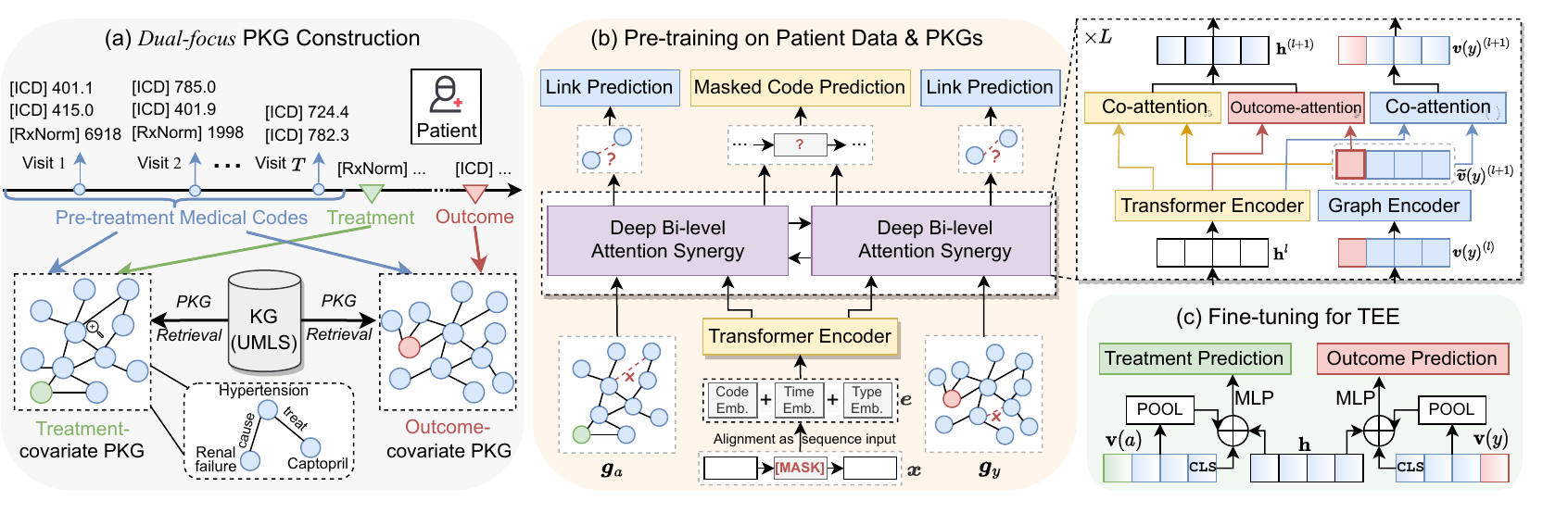}
\caption{A detailed illustration of \model. (a) \textit{Dual-focus} PKGs are constructed by extracting relevant treatment-covariate and outcome-covariate information for each individual patient from KG. (b) The model is pre-trained by synergizing patient data with corresponding PKGs through the proposed deep bi-level attention synergy method. Two pre-training tasks are unified to learn contextualized representations. (c) The pre-trained model is fine-tuned on downstream data for TEE.}
\label{fig:framework}
\end{figure*}

To address this, one might consider deploying foundation models \cite{devlin2018bert,brown2020language,bommasani2021opportunities}
trained on large datasets, to improve generalizability. However, the application of foundation models to TEE is not straightforward. Medical data, often characterized by high-dimensional and sparse medical concepts, continue to pose challenges to these models \cite{huang2019clinicalbert,rasmy2021med}. Even with large-scale datasets, developing a domain-specific understanding of these medical concepts and identifying potential confounders to reduce estimation bias remains difficult. The sheer volume of data does not necessarily equate to rich, specific instances that the model needs to learn effectively.

Therefore, we turn to biomedical knowledge graphs (KGs) - structured representations of diverse medical concepts and their relations. By  synergizing potentially sparse patient data with domain-specific knowledge from KGs, we can derive meaningful insights and identify key confounders for adjustment in TEE. 

Despite its potential, synergizing patient data with KGs poses several challenges. Firstly, utilization of the entire KG can introduce noise that is irrelevant to patient data. While recent works have suggested constructing personalized knowledge graphs (PKGs) by retrieving relevant medical information for each patient from the entire KG to mitigate noise \cite{ye2021medpath, xu2023seqcare, yang2023kerprint}, these methods fail to distinguish among various types of medical codes, such as treatments, covariates, and outcomes. This issue can lead to bias and spurious correlations in TEE. Secondly, the primary application scenario of existing works is clinical risk prediction \cite{choi2017gram, ma2018risk}, neglecting the encoding of vital causal relationships among covariates, treatments, and outcomes unique to TEE. Additionally, existing methods \cite{ye2021medpath, xu2023seqcare} often incorporate patient data and KGs only at the final prediction stage, leading to superficial combination and inefficient information utilization.

To address these challenges, we propose a novel pre-training and fine-tuning framework for TEE, named \model, by synergizing patient data and KGs. Firstly, we address the bias issue by constructing \textit{dual-focus} PKGs, one focusing on the relationship between treatment and covariates (treatment-covariate PKG) and the other focusing on the relationship between outcome and covariates (outcome-covariate PKG). These PKGs explicitly capture and represent the key relationships and dependencies among treatments, outcomes, and covariates, thereby mitigating the risk of spurious correlations. Secondly, we present a deep \textit{bi-level} attention synergy method, namely \texttt{DIVE}. The first level of attention applies a treatment(outcome) attention mechanism to patient data and specific treatment(outcome) information from PKGs, explicitly encoding the complex relationships among covariates, treatments, and outcomes. The second level employs a multi-layer co-attention mechanism to patient data and its corresponding PKGs, ensuring deep information synergy between these modalities.

\model is first pre-trained by combining two self-supervised tasks: masked code prediction and link prediction. These tasks ensure that the patient data and KGs are fully grounded and contextualized to each other. The model is then fine-tuned on downstream data for TEE.

Our contributions include:
\begin{itemize}[leftmargin=*]
\item We propose \model, a novel pre-training and fine-tuning framework that integrates observational patient data with KGs for TEE. This includes the construction of \textit{dual-focus} PKGs and the introduction of a deep bi-level attention synergy method, \texttt{DIVE}.

\item We compile a large dataset of 3M patient records from MarketScan Research Databases\footnote{ \url{https://www.merative.com/real-world-evidence}} and KG (300K nodes, 1M edges) from Unified Medical Language System (UMLS) \cite{bodenreider2004unified} for pre-training, and 4 TEE fine-tuning datasets for assessing treatment effectiveness in coronary artery disease (CAD).

\item Comprehensive experiments demonstrate the superior performance of \model over existing TEE methods. It shows an average improvement of 7\% in AUC for outcome prediction and a 9\% improvement in IF-PEHE for TEE compared to the best baseline across four tasks.

\item Case study shows that the estimated treatment effects align with established randomized controlled trials (RCTs) findings, further demonstrating the effectiveness of our approach in real-world use.
\end{itemize}

\section{Preliminary}
\paragraph{Patient Data.}
A patient record is a collection of multiple visits, denoted as $\overline{\boldsymbol{x}}=\{\boldsymbol{x}_1,\ldots,\boldsymbol{x}_{T}\}$. Each visit is characterized by a series of medications $m_1,\ldots,m_{|\mathcal{M}|}\in\mathcal{M}$ (with $|\mathcal{M}|$ total medication codes), and diagnosis codes $d_1,\ldots,d_{|\mathcal{D}|}\in\mathcal{D}$, (with $|\mathcal{D}|$ total diagnosis codes). A patient's demographics include age and gender, encoded as categorical and binary values respectively, and are denoted as $c\in\mathcal{C}$. We create a comprehensive medical vocabulary, $\mathcal{W}=\{\mathcal{M}, \mathcal{D}, \mathcal{C}\}$ that includes all these patient attributes. We denote the pre-train data as $\mathcal{X}$ and downstream data as $\mathcal{Z}$, where $\mathcal{X}\cap\mathcal{Z}=\varnothing$.

\noindent
\textbf{Personalized Knowledge Graph.} 
A biomedical KG, which contains extensive relationships among various medical codes (e.g., medications and diagnoses), can be represented as a multi-relational graph $\mathcal{G}=(\mathcal{V}, \mathcal{E})$, where $\mathcal{V}$ is the set of entity nodes and $\mathcal{E}\subseteq \mathcal{V}\times\mathcal{R}\times\mathcal{V}$ is the set of edges. These edges connect nodes in $\mathcal{V}$ via triplets, and $\mathcal{R}$ is the set of relation types. A triplet is denoted as ${(h,r,t), h,t\in\mathcal{V}, r\in\mathcal{R}}$, represents a relationship within the KG. As an entire KG can be large and contain noises, a personalized KG (PKG), $g=(v,e)$ is considered by extracting a subgraph from the KG, where $v\in\mathcal{V}$ and $e\in\mathcal{E}$.

\noindent
\textbf{Treatment Effect Estimation.}
Given a patient's visit sequence $\overline{\boldsymbol{x}}=\{\boldsymbol{x}_1,\ldots,\boldsymbol{x}_{T}\}$, demographics $c$, binary treatment $a\in\{0,1\}$ (where 1 indicates treated and 0 indicates control status), disease outcome $y\in\{0,1\}$ (where 1 indicates its presence and 0 indicates its absence), and PKGs $\boldsymbol{g}$, we aim to estimate treatment effect as $\mathbb{E}[Y(1)-Y(0)|X=\overline{\boldsymbol{x}}, C=c, G=\boldsymbol{g}]$, where $Y(A)$ is the potential outcome if the patient receives treatment $A$ \citep{rubin2005causal}. We make three standard assumptions in our TEE analysis: consistency, positivity, and ignorability (see details in Appendix A). These assumptions ensure that the treatment effects estimated as $\mathbb{E}[Y|A=1, X=\overline{\boldsymbol{x}}, C=c, G=\boldsymbol{g}]-\mathbb{E}[Y|A=0, \boldsymbol{X}=\overline{\boldsymbol{x}}, C=c, G=\boldsymbol{g}]$ are identifiable.

\section{Method}
In this section, we introduce our model (Fig. \ref{fig:framework}), including three main modules: 1) \textit{Dual-focus} PKG construction, 2) Pre-training on patient data \& PKGs, and 3) Fine-tuning for TEE. Algorithm \ref{alg:training} shows the model training procedure. 

\subsection{Data Encoding}
As the raw patient data and KGs are not applicable for direct modeling, we need to encode the patient data into dense embeddings and construct \textit{dual-focus} PKGs given the relevant information from individual patients.

\label{sec:pat_data_encoding}

\noindent
\textbf{Patient Data Encoding.}
Compared to natural text, patient data presents unique challenges owing to its irregular temporality (i.e., variability in time intervals between patient visits) and hierarchical structures (i.e., a patient record includes multiple visits and each visit includes different types of medical codes). To address these, we propose a comprehensive embedding approach that extends the original BERT \citep{devlin2018bert} embedding by incorporating both the code type and temporal information. For every input medical code, the patient embedding $\boldsymbol{e}$ is obtained as:
\begin{equation}
\small
\label{eq:pat_data_encoding}
    \boldsymbol{e} =
    \boldsymbol{w}_{\texttt{code}}+
    \boldsymbol{t}_\texttt{type} + \boldsymbol{v}_{\texttt{visit}} + \boldsymbol{p}_{\texttt{physical}}
\end{equation}
where $\boldsymbol{w}_{\texttt{code}}\in\mathbb{R}^{d_\texttt{emb}}$ is the medical code embedding, and $\boldsymbol{t}_\texttt{type}\in\mathbb{R}^{d_\texttt{emb}}$ is the type embedding. Our input data includes three types: demographics, medication, and diagnosis. The visit time embedding $\boldsymbol{v}_{\texttt{visit}}\in\mathbb{R}^{d_\texttt{emb}}$ is the actual time of a visit. The physical time embedding $\boldsymbol{p}_{\texttt{physical}}\in\mathbb{R}^{d_\texttt{emb}}$ measures the physical time over a fixed time interval. 
The code embedding, time embedding, and type embedding are integrated as the input to the patient sequence encoder.

\noindent
\textbf{\textit{Dual-focus} PKG Construction.}
\label{sec:subgraph_retrieval}
To facilitate a nuanced understanding of patient data and obtain personalized estimation for TEE, we propose to construct \textit{dual-focus} PKGs that capture diverse medical contexts of treatment-covariate and outcome-covariate relationship, respectively. We first map the medical codes from a patient's record to their corresponding medical concepts in KG, resulting in an initial set of graph nodes $v'$. Then we augment the graph with treatment and outcome data. The treatment-covariate PKG includes the mapped treatment concepts added to $v'$ as $v'(a)$, while the outcome-covariate PKG incorporates mapped outcome concepts, added to $v'$ as $v'(y)$. To leverage implicit contextual knowledge, we include $k$-hop bridge nodes in the final graph node set $v(a)$ and $v(y)$. A $k$-hop bridge node denotes the entity node positioned within a $k$-hop path between any pair of linked entities in the node set $v'(a)$ or $v'(y)$. Finally, we establish a link between any pair of entities in $v(a)$ and $v(y)$ if there exists an edge between them. This procedure results in \textit{dual-focus} PKGs as $g_a=(v(a), e(a))$ and $g_y=(v(y), e(y))$.

\begin{algorithm}[!t]
\small
\caption{\model Pre-training and Fine-tuning}
\label{alg:training}
\textbf{Input:} Pre-train data $\mathcal{X}$, KG $\mathcal{G}$, downstream data $\mathcal{Z}$ \\
\textbf{Output:} Pre-trained model $f_{\boldsymbol{\theta}^*}$, treatment effects $\delta$

\begin{algorithmic}[1]
\STATE Obtain patient data encoding $\boldsymbol{e}$ by Eq. (\ref{eq:pat_data_encoding});\\
\STATE Extract \textit{dual-focus} PKGs $g(a)$, $g(y)$ from entire KG;\\
\STATE Obtain patient representations $\{\boldsymbol{\tilde{h}}_\texttt{CLS},\boldsymbol{\tilde{h}}_1,\ldots,\boldsymbol{\tilde{h}}_T\}$ by Eq. (\ref{eq:seq_encoder});\\
\STATE Obtain PKG representations $\{\boldsymbol{\tilde{v}}_\texttt{CLS},\boldsymbol{\tilde{v}}_1,\ldots,\boldsymbol{\tilde{v}}_T\}$ by Eq. (\ref{eq:gnn_encoder});\\
\STATE Synergize patient and PKG representations by Eq. (\ref{eq:co-attention});\\
\STATE Pre-train the model by unifying MCP Eq. (\ref{eq:mlm}) and LP Eq. (\ref{eq:loss_link});\\
\STATE Initialize the model with parameters $\boldsymbol{\theta}^*$ from pre-training;\\
\STATE Obtain patient representations $\{\boldsymbol{h}_\texttt{CLS},\boldsymbol{h}_1,\ldots,\boldsymbol{h}_T\}$ and PKG representations $\{\boldsymbol{v}_\texttt{CLS},\boldsymbol{v}_1,\ldots,\boldsymbol{v}_T\}$ by Eq. (\ref{eq:seq_encoder}), (\ref{eq:gnn_encoder});\\
\STATE Fine-tune the model and estimate effects by Eq. (\ref{eq:loss_tee}), (\ref{eq:tee});
\end{algorithmic}
\end{algorithm}

\subsection{Pre-training \model}
The patient embeddings and PKGs are first encoded and then synergized through the proposed deep bi-level attention synergy method. \model is pre-trained by unifying two tasks: masked code prediction and link prediction.
\label{sec:pretrain}

\noindent
\textbf{Patient Sequence Encoder.}
Patient visit sequences are encoded using an N-stacked Transformer \cite{vaswani2017attention}. Each Transformer encoder has a multi-head self-attention layer followed by a fully-connected feed-forward layer. The patient sequence representations are computed as:
\begin{equation}
\small
\label{eq:seq_encoder}
    \boldsymbol{\tilde{h}}^{(l+1)}_\texttt{CLS}, \boldsymbol{\tilde{h}}^{(l+1)}_1,\ldots,\boldsymbol{\tilde{h}}^{(l+1)}_T=f_\texttt{seq}(\boldsymbol{{h}}^{(l)}_\texttt{CLS}, \boldsymbol{{h}}^{(l)}_1,\ldots,\boldsymbol{{h}}^{(l)}_T),
\end{equation}
where $l=1,\ldots, N$ denotes the Transformer layer and the representations in layer $l=0$ are initialized with the patient embedding $\boldsymbol{e}$. The term $\boldsymbol{{h}}_\texttt{CLS}$ is the encoding of a special code that is appended to the patient sequence and acts as the pooling point for prediction. More details of the Transformer architecture are provided in Appendix B.

\noindent
\textbf{Graph Encoder.}
We utilize graph neural networks (GNNs) to encode the PKGs. We initialize node embeddings following existing work \cite{feng2020scalable}. We transform KG triplets into textual data and feed these sentences into a pre-trained language model, BioLinkBERT \cite{yasunaga2022linkbert}, to obtain sentence embeddings. We compute node embeddings by pooling all token outputs of the entity nodes. We encode the treatment-covariate PKG $g(a)$ and outcome-covariate PKG $g(y)$ as follows:
\begin{equation}
\small
\label{eq:gnn_encoder}
\begin{aligned}
    \boldsymbol{\tilde{v}}(a)_\texttt{CLS}^{(l+1)},\ldots,\boldsymbol{\tilde{v}}(a)_I^{(l+1)}&=f_\texttt{gnn}(\boldsymbol{{v}}(a)_\texttt{CLS}^{(l)},\ldots,\boldsymbol{{v}}(a)_I^{(l)}),\\
    \boldsymbol{\tilde{v}}(y)_\texttt{CLS}^{(l+1)},\ldots,\boldsymbol{\tilde{v}}(y)_J^{(l+1)}&=f_\texttt{gnn}(\boldsymbol{{v}}(y)_\texttt{CLS}^{(l)},\ldots,\boldsymbol{{v}}(y)_J^{(l)}),
\end{aligned}
\end{equation}
where $l=1,\ldots,M$ denotes the GNN encoder layer, and $\boldsymbol{{v}}_\texttt{CLS}$ is the encoding of a special node added to the PKG (with edges connecting to all other nodes) to serve as the pooling point for prediction. The node representations are updated via iterative message passing between neighbors as:
\begin{equation}
\small
    \boldsymbol{v}(a)_i^{(l+1)}=f_v(\sum_{s\in\mathcal{N}_i\cup\{i\}}\alpha_{s,i}\boldsymbol{m}_{si})+\boldsymbol{v}(a)_i^{(l)},
\end{equation}
where $\mathcal{N}_i$ denotes the neighbors of entity node $i$, $\alpha_{si}$ is the attention weight for scaling the message pass, and $f_v$ is a multi-layer perceptron with batch normalization. The message $\boldsymbol{m}_{si}$ from $s$ to $i$ is computed as follows:
\begin{equation}
\small
    \boldsymbol{m}_{si}=f_m(\boldsymbol{v}(a)^{(l)}_s, \boldsymbol{r}_{si}),
\end{equation}
where $\boldsymbol{r}_{si}$ is the relation embedding and $f_m$ is a linear transformation. The attention weight $\alpha_{s,i}$, which controls the impact of each neighbor on the current node, is computed as:
\begin{equation}
\small
\begin{aligned}
    \boldsymbol{q}_s=f_q(\boldsymbol{v}(a)_s^{(l)}),\quad&
    \boldsymbol{k}_i=f_k(\boldsymbol{v}(a)_i^{(l)}, \boldsymbol{r}_{si}),\\
    \alpha_{s,i}=\text{Softmax}&(\boldsymbol{q}_s\boldsymbol{k}_i^{\top}/\sqrt{d}),
\end{aligned}
\end{equation} 
where $f_q$ and $f_k$ are linear transformations. The outcome-covariate PKG can be encoded similarly as above.

\noindent
\textbf{Deep Bi-level Attention Synergy.}
To address the challenges of shallow synergy and inefficient information utilization in existing work \cite{ye2021medpath}, we propose a deep bi-level attention synergy method, \texttt{DIVE}. 

The first level of attention handles the complex treatment-covariate and outcome-covariate relationships for bias adjustment and accurate estimation. Given the patient sequence representation $\boldsymbol{\tilde{h}}^{(l)}_{t}$ and treatment node representation $\boldsymbol{\tilde{v}}(a)^{(l)}_{\texttt{a}}$, we compute the treatment attention weight $\alpha_{a,h,t}$ and treatment-related attention-pooling of patient sequence representation $\boldsymbol{\hat{h}}^{(l)}$ as follows:
\begin{equation}
\small
    \label{eq:task-attention}
    \begin{aligned}
        \alpha_{a,h,t}&=\text{Softmax}({\boldsymbol{\tilde{v}}(a)}^{(l)}_{\texttt{a}}{\boldsymbol{\tilde{h}}^{(l)}_{t}}^{\top}/\sqrt{d}),\\
    \hat{h}^{(l)}&=\sum_{t=1}^{T}\alpha_{a,h,t}\boldsymbol{\tilde{h}}_{t}^{(l)}. 
    \end{aligned}
\end{equation}

The second level of attention enables deep synergy of patient data with KGs. We apply a multi-head co-attention \cite{murahari2020large} across patient sequence and graph representations in multiple hidden layers. Concretely, we obtain synergized patient sequence representations by transforming $\boldsymbol{\tilde{h}}^{(l)}$ to queries, $\boldsymbol{\tilde{v}}^{(l)}(a)$ to keys and values. These synergized representations are then concatenated with the treatment-related patient representations derived from Eq. (\ref{eq:task-attention}), and passed through a multi-layer perceptron $f_c$ to yield bi-level attention synergized patient sequence representations $\boldsymbol{h}^{(l)}$. This process is formally denoted as:
\begin{equation}
\small
    \label{eq:co-attention}
    \begin{aligned}
    {\boldsymbol{h}^{(l)}}'&=\texttt{MHCA}_{h,v(a)}(\boldsymbol{\tilde{h}}^{(l)}, \boldsymbol{\tilde{v}}(a)^{(l)}, \boldsymbol{\tilde{v}}(a)^{(l)}),\\
    \boldsymbol{h}^{(l)}&=f_c[{\boldsymbol{h}^{(l)}}'; \boldsymbol{\hat{h}}^{(l)}],\\
    \end{aligned}
\end{equation}
where $\texttt{MHCA}_{Q, K}(Q, K, K)$ is the multi-head co-attention applied to $Q$, $K$, with $Q$ as query,  $K$ as both key and value. We compute the synergized node representations $\boldsymbol{v}(a)^{(l)}$ by transforming $\boldsymbol{\tilde{v}}(a)^{(l)}$ to queries, $\boldsymbol{\tilde{h}}^{(l)}$ to keys and values as:
\begin{equation}
\small
     \boldsymbol{v}(a)=\texttt{MHCA}_{v(a),h}(\boldsymbol{\tilde{v}}(a)^{(l)}, \boldsymbol{\tilde{h}}^{(l)}, \boldsymbol{\tilde{h}}^{(l)}).
\end{equation}
The synergized outcome node representations can be obtained similarly. In essence, \texttt{DIVE} introduces a bi-level attention synergy system: one level that adjusts bias by handling relationships among covariates, treatments, and outcomes, and a second level that focuses on deep synergy. These levels work together to facilitate efficient information utilization and overcome the limitations of shallow synergy methods.

\noindent
\textbf{Pre-training Tasks.}
The goal of pre-training is to encourage a thorough grounding and contextualization of patient data and KGs. To approach this, two self-supervised pre-training tasks are adopted: masked code prediction (MCP) and KG link prediction (LP). 

MCP predicts the masked medical code with position $j\in\mathcal{J}$ using the representation $\boldsymbol{h}_j$. The loss of MCP, $\mathcal{L}_{\text{MCP}}(\theta_m)$, with optimization parameters $\theta_m$, is formulated as:
\begin{equation}
\small
    \mathcal{L}_{\text{MCP}}(\theta^m)=-\sum_{j\in\mathcal{J}}\log({\mathrm{P}(\boldsymbol{w}_j|\boldsymbol{h}_j}),
\label{eq:mlm}
\end{equation}
where $\mathrm{P}(\boldsymbol{w}_j|\boldsymbol{h}_j)$ is the softmax probability of the masked code over all codes in the vocabulary. By using the synergized representations, the patient data are enhanced with external knowledge to predict the masked codes. 

LP is widely used in KG representation learning, which aims to distinguish existing (positive) triplets from corrupted (negative) triplets using the representations of entities and relations. Formally, given the representations of a triplet obtained from the graph encoder as $(\boldsymbol{v}_h, \boldsymbol{r},\boldsymbol{v}_t)$, the pre-training loss for LP is optimized over parameters $\theta^l$ as:
\begin{equation}
\small
\label{eq:loss_link}
\begin{aligned}
        \mathcal{L}_\text{LP}(\theta^l)=\sum_{(h,r,t)\in S}\Big(-\sigma(d(\boldsymbol{v}_h,\boldsymbol{r},\boldsymbol{v}_t)) + \\
    \sum_{(h',r,t')\in S'}\sigma(d(\boldsymbol{v}_{h'},\boldsymbol{r},\boldsymbol{v}_{t'}))\Big)
\end{aligned}
\end{equation}
where $(h',r,t')\in S'_{(h,r,t)}$ are corrupted triplets in which either the head or tail entity is replaced by a random entity (but not both simultaneously), $\sigma$ denotes logarithm sigmoid function, and $d$ is the score function such as TransE \cite{bordes2013translating} and DistMult \cite{yang2014embedding}. 

The final pre-training loss is optimized over parameters $\boldsymbol{\theta}=\{\theta^m, \theta^l\}$ by integrating MLP and LP as $\mathcal{L}(\boldsymbol{\theta})=\mathcal{L}_\text{MCP}(\theta^m)+\mathcal{L}_\text{LP}(\theta^l)$.

\subsection{Fine-tuning \model for TEE}
\label{sec:finetune}
After pre-training, we fine-tune the model on downstream data for TEE. To mitigate confounding bias, the model is fine-tuned to simultaneously predict the treatment and outcome using shared representations. This strategy discourages reliance on unrelated features and prioritizes confounders for predictions \cite{shi2019adapting}.

To elaborate, given downstream patient data and corresponding PKGs, 
we obtain representations for patients, treatment-covariate PKG, and outcome-covariate PKG.
We then predict the treatment using a combination of  $\boldsymbol{h}_\texttt{CLS}$, $\boldsymbol{v}(a)_\texttt{CLS}$,
and an attention-based pooling $\boldsymbol{v}(a)_\texttt{POOL}$ with query $\boldsymbol{h}_\texttt{[CLS]}$
as the input to a prediction head $f_{\phi^a}$. The loss of treatment prediction is computed as:
\begin{equation}
\small
\begin{aligned}
    \hat{a}=f_{\phi^a}\circ f_{\boldsymbol{\theta}^*}([&\boldsymbol{h}_\texttt{CLS},\boldsymbol{v}(a)_\texttt{CLS},\boldsymbol{v}(a)_\texttt{POOL}]),\\
    \mathcal{L}_{\text{T}}(\boldsymbol{\theta}^*,\phi^a)&=\text{BCE}(\hat{a}, a),
\end{aligned}
\end{equation}
where $\boldsymbol{\theta}^*$ are the optimized parameters of the pre-train model, \text{BCE} denotes binary cross entropy loss. Similarly, we predict the outcome by combining $\boldsymbol{h}_\texttt{CLS}$, $\boldsymbol{v}(y)_\texttt{CLS}$,
and an attention-based pooling $\boldsymbol{v}(y)_\texttt{POOL}$ with query $\boldsymbol{h}_\texttt{[CLS]}$
as the input to a prediction head $f_{\phi^y}$. We employ separate heads for treated and control potential outcomes, computing the loss of outcome prediction as:
\begin{equation}
\small
\begin{aligned}
    \hat{y}=f_{\phi^y}\circ f_{\boldsymbol{\theta}^*}([&\boldsymbol{h}_\texttt{CLS},\boldsymbol{v}(y)_\texttt{CLS},\boldsymbol{v}(y)_\texttt{POOL}]),\\
    \mathcal{L}_{\text{O}}(\boldsymbol{\theta}^*,\phi^y)&=\text{BCE}(\hat{y}, y).
\end{aligned}
\end{equation}

We jointly optimize both treatment prediction and outcome prediction, computing the final loss as follows:
\begin{equation}
\small
\label{eq:loss_tee}
    \mathcal{L}_\text{TEE}(\boldsymbol{\theta}^*,\boldsymbol{\phi})=\mathcal{L}_{\text{O}}(\boldsymbol{\theta}^*,\phi^y)+\beta\mathcal{L}_{\text{T}}(\boldsymbol{\theta}^*,\phi^a),
\end{equation}
where $\beta$ is a hyper-parameter that controls the influence of treatment prediction. Note that only the observational, or factual, outcomes are used to compute outcome prediction loss, as counterfactual outcomes are unavailable.
After fine-tuning the model, we infer the treatment effect $\delta$ 
as the difference between two predicted potential outcomes under the treated and control treatment as:
\begin{equation}
\small
\label{eq:tee}
\hat{\delta}=\hat{y}_{a=1} - \hat{y}_{a=0} 
\end{equation}

\section{Experimental Setup}
\label{sec:exp}
\subsubsection{Pre-training Data.} We extract patient data from MarketScan Commercial Claims and Encounters (CCAE) database\footnote{https://www.merative.com/real-world-evidence} for those diagnosed with coronary artery disease (CAD), producing a dataset of 2,955,399 patient records and 116,661 unique medical codes. We use UMLS\footnote{\url{https://www.nlm.nih.gov/research/umls/index.html}} as external knowledge. We unify the medical codes in patient data and concepts in the UMLS
through standard vocabularies\footnote{\url{nlm.nih.gov/research/umls/sourcereleasedocs/index.html}} and extract all relevant relationships from UMLS, resulting in a KG with around 300K nodes and 1M edges

\noindent
\textbf{Downstream Tasks.} 
Our goal is to evaluate the effects of two treatments on reducing stroke and myocardial infarction risk for CAD patients, given the patient's covariates and corresponding PKGs. As the ground truth treatment effects are not available in observational data and RCTs are the gold standard for TEE, we specifically create 4 downstream datasets based on CAD-related RCTs. More details of datasets are provided in Appendix Table \ref{tab:downstream_stat} and Fig. \ref{fig:downstream_data}).

\noindent
\textbf{Baselines.} We compare \model with state-of-the-art methods, all trained solely on downstream data. 

\begin{itemize}[leftmargin=*]
     \item \textbf{TARNet} \citep{shalit2017estimating} 
    predicts the potential outcomes based on balanced hidden representations among treated and controlled groups.  
    \item
     \textbf{DragonNet} \citep{shi2019adapting} jointly predicts treatment and outcome via a three-head neural network: one for treatment prediction and two for outcomes. 
    \item 
    \textbf{DR-CFR} \citep{hassanpour2019learning} predicts the counterfactual outcome by learning disentangled representations that the covariates can be disentangled into three components: only contributing to treatment selection, only contributing to outcome predication, and both.
    \item 
    \textbf{TNet} \citep{curth2021nonparametric} is a deep neural network version of T-learner 
    \citep{kunzel2019metalearners} (i.e., decomposes the TEE into two or more sub-regression/classification problems).  
    \item 
    \textbf{SNet} \citep{curth2021nonparametric} learns disentangled representations and assumes that the covariates can be disentangled into five components by considering two potential outcomes separately. 
    \item 
    \textbf{FlexTENet} \citep{curth2021inductive} assumes inductive bias for the shared structure of two potential outcomes and adaptively learns what to share between the potential outcome functions.
    \item 
    \textbf{TransTEE} \citep{zhang2022can} is a Transformer-based TEE model, which encodes the covariates and treatments via Transformer and cross-attention. 
\end{itemize}

Additionally, we consider several variants of the proposed model for comparison:
\begin{itemize}[leftmargin=*]
    \item \textit{w/o} \texttt{DIVE}: replacing the proposed deep bi-level attention synergy method \texttt{DIVE} with a simple concatenation of patient and graph representations only at the final prediction.
    
    \item \textit{w/o KGs}: pre-trained solely on patient data without KGs.

    \item \textit{w/o pre-train}: directly trained on the downstream patient data and KGs using the same model architecture as \model.

    \item \textit{w/o pre-train \& KGs}: directly trained on the downstream patient data with the patient sequence encoder only.

\end{itemize}

Note that we do not compare with existing pre-training models for clinical risk prediction \cite{li2020behrt,rasmy2021med}, as these models are not directly applicable to our context. Clinical risk prediction models focus on forecasting the likelihood of a disease based on variable correlations. TEE mainly quantifies the causal impact of a treatment on an outcome, predicting all potential outcomes as treatment effects, an entirely different objective from risk prediction.

\noindent
\textbf{Metrics.} We assess factual prediction performance using standard classification metrics: Area under the ROC Curve (AUC) and Area under the Precision-Recall Curve (AUPR). We evaluate counterfactual prediction performance using influence function-based precision of estimating heterogeneous effects (IF-PEHE) \citep{alaa2019validating}, which measures the mean squared error between estimated treatment effects and approximated true treatment effects. 
Additional details of this metric are in Appendix C.

\noindent
\textbf{Implementation Details.}
The patient sequence encoder uses the BERT-base architecture \citep{devlin2018bert}. The PKGs are retrieved with 2-hop bridge nodes, with a maximum node limit of 200. The downstream data is randomly split into training, validation, and test sets with percentages of 90\%, 5\%, and 5\% respectively. All results are reported on the test sets. More implementation details\footnote{Code: \url{https://github.com/ruoqi-liu/KG-TREAT}} including parameter tuning and setup are mentioned in Appendix C.

\section{Results}

\begin{table*}[!th]
\centering
\begin{tabular}
{@{\extracolsep{\fill}} l|ccc|ccc}
\toprule
\multirow{2}{*}{Method} & \multicolumn{3}{c|}{Rivaroxaban v.s. Aspirin} & \multicolumn{3}{c}{Valsartan v.s. Ramipril} \\ \cmidrule(l){2-7} 
 & AUC $\uparrow$ & AUPR $\uparrow$ & IF-PEHE $\downarrow$ & AUC $\uparrow$ & AUPR $\uparrow$ & IF-PEHE $\downarrow$  \\ \midrule
TARNet & 0.746 & 0.404 & 0.277 & 0.743 & 0.333 & 0.289  \\
DragonNet & 0.761 & 0.424 & 0.261 & 0.742 & 0.337 & 0.271  \\
DR-CFR & 0.764 & 0.426 & 0.259 & 0.747 & 0.341 & 0.276  \\
TNet & 0.726 & 0.401 & 0.321 & 0.730 & 0.322 & 0.299  \\
SNet & 0.761 & 0.430 & 0.254 & 0.747 & 0.354 & 0.268  \\
FlexTENet & 0.729 & 0.403 & 0.301 & 0.739 & 0.328 & 0.285 \\
TransTEE & 0.751 & 0.411 & 0.272 & 0.753 & 0.379 & 0.264  \\ \midrule
\model & $\mathbf{0.828}$ & $\mathbf{0.556}$ & $\mathbf{0.171}$ &
$\mathbf{0.858}$ & $\mathbf{0.526}$ & $\mathbf{0.149}$  \\
\;  \textit{w/o} \texttt{DIVE} & 0.811 & 0.522 & 0.202 & 0.829 & 0.495 & 0.160  \\
\;  \textit{w/o KGs} & 0.805 & 0.518 & 0.219 & 0.813 & 0.481 & 0.165  \\
\;  \textit{w/o pre-train} & 0.786 & 0.488 & 0.231 & 0.778 & 0.373 & 0.189  \\ 
\;  \textit{w/o pre-train \& KGs} & 0.769 & 0.470 & 0.239 & 0.749 & 0.371 & 0.198  \\ 
\midrule
\multirow{2}{*}{Method} & \multicolumn{3}{c|}{Ticagrelor v.s.   Aspirin} & \multicolumn{3}{c}{Apixaban v.s. Warfarin} \\ \cmidrule(l){2-7} 
 & AUC $\uparrow$ & AUPR $\uparrow$ & IF-PEHE $\downarrow$ & AUC $\uparrow$ & AUPR $\uparrow$ & IF-PEHE $\downarrow$  \\ \midrule
TARNet & 0.755 & 0.460 & 0.282 & 0.760 & 0.515 & 0.325 \\
DragonNet & 0.762 & 0.464 & 0.289 & 0.766 & 0.534 & 0.309 \\
DR-CFR & 0.764 & 0.460 & 0.278 & 0.769 & 0.526 & 0.287 \\
TNet & 0.741 & 0.433 & 0.311 & 0.753 & 0.509 & 0.330 \\
SNet & 0.765 & 0.465 & 0.265 & 0.769 & 0.527 & 0.273 \\
FlexTENet & 0.750 & 0.452 & 0.313 & 0.756 & 0.512 & 0.324 \\
TransTEE & 0.770 & 0.471 & 0.255 & 0.803 & 0.534 & 0.267 \\ \midrule
\model & $\mathbf{0.851}$ & $\mathbf{0.609}$ & 
$\mathbf{0.160}$ &
$\mathbf{0.843}$ & $\mathbf{0.639}$ & 
$\mathbf{0.191}$ \\
\;  \textit{w/o} \texttt{DIVE} & 0.839 & 0.571 & 0.176 & 0.830 & 0.611 & 0.210 \\
\;  \textit{w/o KGs} & 0.830 & 0.552 & 0.181 & 0.829 & 0.605 & 0.217 \\
\;  \textit{w/o pre-train} & 0.815 & 0.524 & 0.200 & 0.807 & 0.569 & 0.241 \\ 
\;  \textit{w/o pre-train \& KGs} & 0.808 & 0.497 & 0.213 & 0.801 & 0.550 & 0.247 \\ 
 \bottomrule
\end{tabular}%
\caption{Comparison with state-of-the-art methods on 4 downstream datasets. \texttt{DIVE} is our proposed deep bi-level attention method for synergizing patient data with KGs. The results are averaged over 20 random runs.}
\label{tab:comparison_sota_brief}
\end{table*}
\begin{table*}[!th]
\centering
\adjustbox{max width=\textwidth}{%
\begin{tabular}{lcccc}
\toprule
Target v.s. Compared
& Estimated Effect 
& P value 
& Model Conclusion
& RCT Conclusion
\\ 
\midrule
Rivaroxaban v.s. Aspirin 
& {[}-0.010, 0.009{]}
& 0.952
& No significant difference
& No significant difference \citep{anand2018rivaroxaban}          
\\
Valsartan v.s. Ramipril  
& {[}-0.015, 0.007{]}  
& 0.564
& No significant difference
& No significant difference \citep{pfeffer2021angiotensin}         
\\
Ticagrelor v.s. Aspirin  
& {[}-0.006, 0.021{]} 
& 0.436
& No significant difference
& No significant difference \citep{sandner2020ticagrelor}        
\\
Apixaban v.s. Warfarin   
& {[}-0.006, -0.001{]}   
& 0.001
& A. is more effective than W.
& A. is more effective than W. \citep{granger2011apixaban}
\\ \bottomrule
\end{tabular}}
\caption{Comparison of the estimated treatment effects with corresponding ground truth RCT. The estimated effects are shown in 95\% confidence intervals (CI) under 20 bootstrap runs. The RCT conclusions are obtained from published articles.}
\label{tab:rct_validation}
\end{table*}

\subsection{Quantitative Analysis}
We quantitatively compare \model with state-of-the-art methods in terms of factual outcome prediction and TEE. Table \ref{tab:comparison_sota_brief} presents the results on 4 downstream datasets. Our key findings include:
\begin{itemize}[leftmargin=*]
    \item \model significantly outperforms the best baseline method, demonstrating an average improvement of 7\% in AUC, 12\% in AUPR, and 9\% in IF-PEHE. This validates the effectiveness of our pre-training approach, which synergizes patient data with KGs.

    \item The variant of \model without the deep bi-level attention synergy method, \textit{w/o} \texttt{DIVE}, shows a drop in performance. This highlights the effectiveness of \texttt{DIVE} in modeling the relationships among covariates, treatments, and outcomes, and in synergizing patient data with KGs.

    \item Pre-training has a more significant impact on model performance than KGs, as indicated by the greater performance decline in the \textit{w/o pre-train} scenario compared to the \textit{w/o KGs} scenario. And \textit{w/o pre-train \& KGs} yields the worst performance among all the model variants.
\end{itemize}

\subsection{Qualitative Analysis}
Besides the quantitative analysis, we demonstrate the model's ability in facilitating randomized controlled trials (RCTs) by providing an accurate estimation of treatment effects and generating consistent conclusions. Additionally, we show that KGs help improve performance by identifying a more comprehensive set of confounders for adjustment.

\begin{figure*}[!th]
\centering
\includegraphics[width=0.85\linewidth]{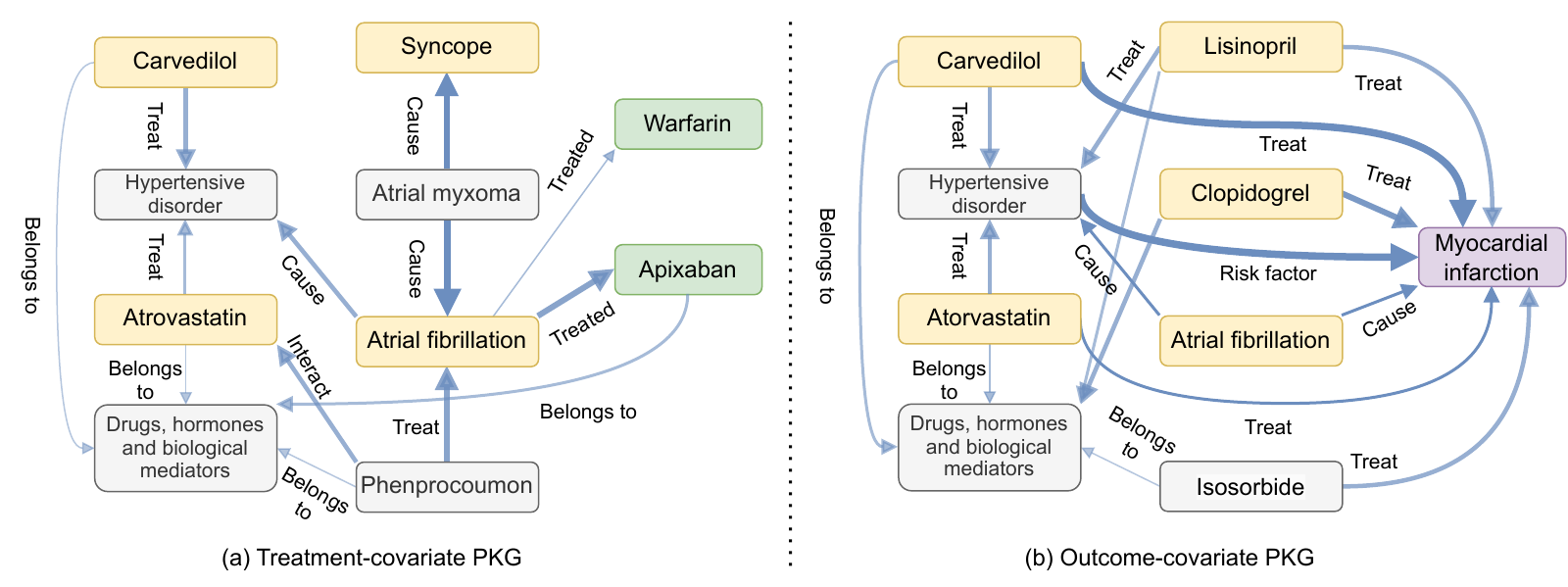}
\caption{Visualization of the graph attention weights for (a) treatment-covariate PKG and (b) outcome-covariate PKG. The patient is from ``Apixaban v.s. Warfarin'' dataset. Higher attention weights are denoted as thicker and darker edges in the graph. The extra 2-hop bridge nodes are in gray color to distinguish them from the initial set of nodes in yellow color.}
\label{fig:att_vis}
\end{figure*}

\noindent
\textbf{Validation with RCT Conclusion.}
We compare the estimated treatment effects to the corresponding RCT results. First, we compute the average treatment effects as the mean of the differences between the treated outcomes and controlled outcomes \citep{hernan2004definition}. Then, we test the significance of the estimated effects against zero using a T-test with significance level $\alpha=0.05$ for model conclusion generation. Table \ref{tab:rct_validation} shows that model-generated conclusions align with their corresponding RCT conclusions, suggesting that \model can be served as an effective computational tool to emulate RCTs using large-scale patient data and KGs. A thorough comparison of our method with all the baseline methods is shown in Appendix Table \ref{tab:rct_all}.

\begin{table*}[!th]
\centering
\small
\begin{tabular}{ll|cc|cc|cc|cc}
\toprule
&
  \multirow{2}{*}{Method} &
  \multicolumn{2}{c|}{Rivaroxaban v.s. Aspirin} &
  \multicolumn{2}{c|}{Valsartan v.s. Ramipril} &
  \multicolumn{2}{c|}{Ticagrelor v.s. Aspirin} &
  \multicolumn{2}{c}{Apixaban v.s. Warfarin} \\
\multicolumn{2}{l|}{} &
  AUC $\uparrow$ &
  IF-PEHE $\downarrow$ &
  AUC $\uparrow$ &
  IF-PEHE $\downarrow$ &
  AUC $\uparrow$ &
  IF-PEHE $\downarrow$ &
  AUC $\uparrow$ &
  IF-PEHE $\downarrow$ \\ \midrule
\multirow{2}{*}{\begin{tabular}[c]{@{}l@{}}Pre-train\\ Task\end{tabular}} & MCP only & 0.815 & 0.204 & 0.840 & 0.160 & 0.840 & 0.169 & 0.831 & 0.200 \\
& LP only  & 0.791 & 0.225 & 0.813 & 0.174 & 0.816 & 0.186 & 0.812 & 0.232 \\ \midrule
\multirow{2}{*}{\begin{tabular}[c]{@{}l@{}}Score\\ Function\end{tabular}} & RotatE   & 0.825 & 0.176 & 0.851 & 0.152 & 0.848 & 0.164 & 0.838 & 0.198 \\
& TransE   & 0.823 & 0.177 & 0.851 & 0.155 & 0.846 & 0.166 & 0.832 & 0.203 \\ \midrule
 &
  \model &
  $\mathbf{0.828}$ &
  $\mathbf{0.171}$ &
  $\mathbf{0.858}$ &
  $\mathbf{0.149}$ &
  $\mathbf{0.851}$ &
  $\mathbf{0.160}$ &
  $\mathbf{0.843}$ &
  $\mathbf{0.191}$ \\ \bottomrule
\end{tabular}
\caption{Ablation study results of using different pre-train tasks and score functions in link prediction. \model adopts both MCP and LP as pre-training tasks and DistMult as score function.}
\label{tab:ablation_all}
\end{table*}

\noindent
\textbf{Attention Visualization.} 
We use case studies to illustrate how our model identifies potential confounders from patient data and KGs for adjusting bias and accurate estimation. We visualize the model attention weights of each PKG in Fig. \ref{fig:att_vis} and observe that \model successfully identifies key confounders as medical codes with high attention weights from PKGs. 
For example, the common medical codes, such as  ``Carvedilol'', ``Hypertensive
disorder'', ``Atorvastatin'', are identified as the potential confounders and also mentioned in related literature \cite{stolk2017risk, yusuf2004effect}. 
Notably, with the help of external knowledge, our model can recover potential confounders that are not observed in the patient data. For example, ``Hypertensive disorder'' is a potentially missing confounding factor added to PKG through 2-hop bridge node searching. This result indicates that solely relying on the patient data may fail to recognize a more comprehensive set of confounders for adjustment.  

\subsection{Ablation Studies}
In the above experiments, we find that pre-training plays a critical role in model performance. Therefore, we focus on important model choices in pre-training tasks. An additional ablation study of the influence of downstream data size on model performance is provided in Appendix Fig. \ref{fig:ablation_few_shot}.

\noindent
\textbf{Pre-training Tasks.}
We analyze the impact of pre-training tasks by excluding the LP and MCP tasks. As shown in Table \ref{tab:ablation_all}, unifying both MCP and LP tasks (\model) yields the best performance on 4 downstream datasets. The MCP task plays a crucial role in pre-training, with its exclusion leading to larger performance drops (3\% of AUC and 4\% of IF-PEHE) than the LP task (1\% of AUC and IF-PEHE). This demonstrates the importance of integrating both tasks to jointly learn from both data modalities.

\noindent
\textbf{Link Prediction Head Choice.}
We evaluate various score functions for the LP task. As shown in Table \ref{tab:ablation_all}, DistMult, which is adopted in our model, offers better performance than other score functions. While, the differences among different functions are not significant, indicating that the pre-training model on large-scale data is not particularly sensitive to the scoring function selection.

\section{Related Work}
\textbf{Deep learning for TEE.} 
Deep learning has been extensively used for TEE and achieved improved performance than classical linear methods due to its flexibility of modeling non-linearity \citep{shalit2017estimating,shi2019adapting,hassanpour2019learning,curth2021nonparametric,curth2021inductive}. For instance, TARNet \citep{shalit2017estimating} employs shared representations to simultaneously predict two potential outcomes, encouraging the similarity between treated and control distributions. SNet \citep{curth2021nonparametric} learns disentangled representations for flexible information sharing among treatment prediction and outcome prediction. Recent Transformer-based models have been introduced as backbones for TEE to help handle various data modalities (e.g., graph, text, etc.) \citep{zhang2022can, guo2021cetransformer}. However, these existing methods are mainly trained on small-scale, task-specific labeled data. This may limit model performance due to insufficient learning of the complex relationships among covariates, treatments, and outcomes.

\noindent
\textbf{Knowledge Integration in Healthcare.}
Various works in healthcare incorporate biomedical KGs to enrich patient data \cite{choi2017gram, ma2018risk, ma2018kame}. Recent works have shown that personalized KGs (PKGs), constructed by retrieving relevant medical features of individual patients from KG, can encode personalized information and mitigate noise of the entire KG \cite{ye2021medpath, xu2023seqcare, yang2023kerprint}. However, these approaches often fail to distinguish among different types of medical codes and encode relationships among covariates, treatments, and outcomes. This can lead to confounding bias and spurious correlations in TEE. Additionally, existing methods often integrate patient data and KGs only in the final prediction, resulting in shallow combination and inefficient information utilization. 

\noindent
\textbf{Pre-training in Healthcare.}
The foundation models have been successfully applied to various domains including healthcare and patient data \cite{huang2019clinicalbert,li2020behrt, rasmy2021med}. These methods typically convert a patient's medical records into a sequence of tokens for pre-training, followed by fine-tuning for healthcare-related downstream tasks such as clinical risk prediction. However, learning deep, domain-specific representations of complex medical features from sparse patient data can be challenging, even with large-scale data. Furthermore, existing methods often fail to handle complex relationships among covariates, treatments, and outcomes, potentially leading to biased estimation. To the best of our knowledge, ours is the first pre-training model by synergizing both observational patient data and KGs.

\section{Conclusion}
In this paper, we propose \model, a pre-training and fine-tuning framework for TEE by synergizing patient data with KGs. We construct \textit{dual-focus} personalized KGs that incorporate key relationships among covariates, treatments, and outcomes for addressing potential bias in TEE. We propose a novel synergy method (\texttt{DIVE}) to achieve deep information exchange between patient data and PKGs, and encourage complex relationship encoding for TEE. We jointly pre-train the model via two self-supervised tasks and fine-tune it on downstream TEE datasets. Thorough experiments on real-world patient data show the effectiveness of \model compared to state-of-the-art methods. We further demonstrate the estimated treatment effects are well consistent with corresponding published RCTs. 

\section{Acknowledgments}
This work was funded in part by the National Institute of General Medical Sciences (NIGMS) of NIH under award number R01GM141279.


\begin{thebibliography}{38}
\providecommand{\natexlab}[1]{#1}

\bibitem[{Alaa and van~der Schaar(2019)}]{alaa2019validating}
Alaa, A.~M.; and van~der Schaar, M. 2019.
\newblock Validating Causal Inference Models via Influence Functions.
\newblock In Chaudhuri, K.; and Salakhutdinov, R., eds., \emph{Proceedings of the 36th International Conference on Machine Learning, {ICML} 2019, 9-15 June 2019, Long Beach, California, {USA}}, volume~97 of \emph{Proceedings of Machine Learning Research}, 191--201.

\bibitem[{Anand et~al.(2018)Anand, Bosch, Eikelboom, Connolly, Diaz, Widimsky, Aboyans, Alings, Kakkar, Keltai et~al.}]{anand2018rivaroxaban}
Anand, S.~S.; Bosch, J.; Eikelboom, J.~W.; Connolly, S.~J.; Diaz, R.; Widimsky, P.; Aboyans, V.; Alings, M.; Kakkar, A.~K.; Keltai, K.; et~al. 2018.
\newblock Rivaroxaban with or without aspirin in patients with stable peripheral or carotid artery disease: an international, randomised, double-blind, placebo-controlled trial.
\newblock \emph{The Lancet}, 391(10117): 219--229.

\bibitem[{Bodenreider(2004)}]{bodenreider2004unified}
Bodenreider, O. 2004.
\newblock The unified medical language system (UMLS): integrating biomedical terminology.
\newblock \emph{Nucleic acids research}, 32(suppl\_1): D267--D270.

\bibitem[{Bommasani et~al.(2021)Bommasani, Hudson, Adeli, Altman, Arora, von Arx, Bernstein, Bohg, Bosselut, Brunskill et~al.}]{bommasani2021opportunities}
Bommasani, R.; Hudson, D.~A.; Adeli, E.; Altman, R.; Arora, S.; von Arx, S.; Bernstein, M.~S.; Bohg, J.; Bosselut, A.; Brunskill, E.; et~al. 2021.
\newblock On the opportunities and risks of foundation models.
\newblock \emph{ArXiv preprint}, abs/2108.07258.

\bibitem[{Bordes et~al.(2013)Bordes, Usunier, Garc{\'{\i}}a{-}Dur{\'{a}}n, Weston, and Yakhnenko}]{bordes2013translating}
Bordes, A.; Usunier, N.; Garc{\'{\i}}a{-}Dur{\'{a}}n, A.; Weston, J.; and Yakhnenko, O. 2013.
\newblock Translating Embeddings for Modeling Multi-relational Data.
\newblock In Burges, C. J.~C.; Bottou, L.; Ghahramani, Z.; and Weinberger, K.~Q., eds., \emph{Advances in Neural Information Processing Systems 26: 27th Annual Conference on Neural Information Processing Systems 2013. Proceedings of a meeting held December 5-8, 2013, Lake Tahoe, Nevada, United States}, 2787--2795.

\bibitem[{Brown et~al.(2020)Brown, Mann, Ryder, Subbiah, Kaplan, Dhariwal, Neelakantan, Shyam, Sastry, Askell, Agarwal, Herbert{-}Voss, Krueger, Henighan, Child, Ramesh, Ziegler, Wu, Winter, Hesse, Chen, Sigler, Litwin, Gray, Chess, Clark, Berner, McCandlish, Radford, Sutskever, and Amodei}]{brown2020language}
Brown, T.~B.; Mann, B.; Ryder, N.; Subbiah, M.; Kaplan, J.; Dhariwal, P.; Neelakantan, A.; Shyam, P.; Sastry, G.; Askell, A.; Agarwal, S.; Herbert{-}Voss, A.; Krueger, G.; Henighan, T.; Child, R.; Ramesh, A.; Ziegler, D.~M.; Wu, J.; Winter, C.; Hesse, C.; Chen, M.; Sigler, E.; Litwin, M.; Gray, S.; Chess, B.; Clark, J.; Berner, C.; McCandlish, S.; Radford, A.; Sutskever, I.; and Amodei, D. 2020.
\newblock Language Models are Few-Shot Learners.
\newblock In Larochelle, H.; Ranzato, M.; Hadsell, R.; Balcan, M.; and Lin, H., eds., \emph{Advances in Neural Information Processing Systems 33: Annual Conference on Neural Information Processing Systems 2020, NeurIPS 2020, December 6-12, 2020, virtual}.

\bibitem[{Chen and Guestrin(2016)}]{chen2016xgboost}
Chen, T.; and Guestrin, C. 2016.
\newblock XGBoost: {A} Scalable Tree Boosting System.
\newblock In Krishnapuram, B.; Shah, M.; Smola, A.~J.; Aggarwal, C.~C.; Shen, D.; and Rastogi, R., eds., \emph{Proceedings of the 22nd {ACM} {SIGKDD} International Conference on Knowledge Discovery and Data Mining, San Francisco, CA, USA, August 13-17, 2016}, 785--794.

\bibitem[{Choi et~al.(2017)Choi, Bahadori, Song, Stewart, and Sun}]{choi2017gram}
Choi, E.; Bahadori, M.~T.; Song, L.; Stewart, W.~F.; and Sun, J. 2017.
\newblock {GRAM:} Graph-based Attention Model for Healthcare Representation Learning.
\newblock In \emph{Proceedings of the 23rd {ACM} {SIGKDD} International Conference on Knowledge Discovery and Data Mining, Halifax, NS, Canada, August 13 - 17, 2017}, 787--795.

\bibitem[{Curth and van~der Schaar(2021{\natexlab{a}})}]{curth2021nonparametric}
Curth, A.; and van~der Schaar, M. 2021{\natexlab{a}}.
\newblock Nonparametric Estimation of Heterogeneous Treatment Effects: From Theory to Learning Algorithms.
\newblock In Banerjee, A.; and Fukumizu, K., eds., \emph{The 24th International Conference on Artificial Intelligence and Statistics, {AISTATS} 2021, April 13-15, 2021, Virtual Event}, volume 130 of \emph{Proceedings of Machine Learning Research}, 1810--1818.

\bibitem[{Curth and van~der Schaar(2021{\natexlab{b}})}]{curth2021inductive}
Curth, A.; and van~der Schaar, M. 2021{\natexlab{b}}.
\newblock On Inductive Biases for Heterogeneous Treatment Effect Estimation.
\newblock In Ranzato, M.; Beygelzimer, A.; Dauphin, Y.~N.; Liang, P.; and Vaughan, J.~W., eds., \emph{Advances in Neural Information Processing Systems 34: Annual Conference on Neural Information Processing Systems 2021, NeurIPS 2021, December 6-14, 2021, virtual}, 15883--15894.

\bibitem[{Devlin et~al.(2019)Devlin, Chang, Lee, and Toutanova}]{devlin2018bert}
Devlin, J.; Chang, M.-W.; Lee, K.; and Toutanova, K. 2019.
\newblock {BERT}: Pre-training of Deep Bidirectional Transformers for Language Understanding.
\newblock In \emph{Proceedings of the 2019 Conference of the North {A}merican Chapter of the Association for Computational Linguistics: Human Language Technologies, Volume 1 (Long and Short Papers)}, 4171--4186.

\bibitem[{Feng et~al.(2020)Feng, Chen, Lin, Wang, Yan, and Ren}]{feng2020scalable}
Feng, Y.; Chen, X.; Lin, B.~Y.; Wang, P.; Yan, J.; and Ren, X. 2020.
\newblock Scalable Multi-Hop Relational Reasoning for Knowledge-Aware Question Answering.
\newblock In \emph{Proceedings of the 2020 Conference on Empirical Methods in Natural Language Processing (EMNLP)}, 1295--1309.

\bibitem[{Glass et~al.(2013)Glass, Goodman, Hern{\'a}n, and Samet}]{glass2013causal}
Glass, T.~A.; Goodman, S.~N.; Hern{\'a}n, M.~A.; and Samet, J.~M. 2013.
\newblock Causal inference in public health.
\newblock \emph{Annual review of public health}, 34: 61--75.

\bibitem[{Granger et~al.(2011)Granger, Alexander, McMurray, Lopes, Hylek, Hanna, Al-Khalidi, Ansell, Atar, Avezum et~al.}]{granger2011apixaban}
Granger, C.~B.; Alexander, J.~H.; McMurray, J.~J.; Lopes, R.~D.; Hylek, E.~M.; Hanna, M.; Al-Khalidi, H.~R.; Ansell, J.; Atar, D.; Avezum, A.; et~al. 2011.
\newblock Apixaban versus warfarin in patients with atrial fibrillation.
\newblock \emph{New England Journal of Medicine}, 365(11): 981--992.

\bibitem[{Guo et~al.(2021)Guo, Zheng, Liu, Yan, and Zhu}]{guo2021cetransformer}
Guo, Z.; Zheng, S.; Liu, Z.; Yan, K.; and Zhu, Z. 2021.
\newblock CETransformer: Casual Effect Estimation via Transformer Based Representation Learning.
\newblock In \emph{Chinese Conference on Pattern Recognition and Computer Vision (PRCV)}, 524--535. Springer.

\bibitem[{Hassanpour and Greiner(2020)}]{hassanpour2019learning}
Hassanpour, N.; and Greiner, R. 2020.
\newblock Learning Disentangled Representations for CounterFactual Regression.
\newblock In \emph{8th International Conference on Learning Representations, {ICLR} 2020, Addis Ababa, Ethiopia, April 26-30, 2020}.

\bibitem[{Hern{\'a}n(2004)}]{hernan2004definition}
Hern{\'a}n, M.~A. 2004.
\newblock A definition of causal effect for epidemiological research.
\newblock \emph{Journal of Epidemiology \& Community Health}, 58(4): 265--271.

\bibitem[{Huang, Altosaar, and Ranganath(2019)}]{huang2019clinicalbert}
Huang, K.; Altosaar, J.; and Ranganath, R. 2019.
\newblock Clinicalbert: Modeling clinical notes and predicting hospital readmission.
\newblock \emph{ArXiv preprint}, abs/1904.05342.

\bibitem[{K{\"u}nzel et~al.(2019)K{\"u}nzel, Sekhon, Bickel, and Yu}]{kunzel2019metalearners}
K{\"u}nzel, S.~R.; Sekhon, J.~S.; Bickel, P.~J.; and Yu, B. 2019.
\newblock Metalearners for estimating heterogeneous treatment effects using machine learning.
\newblock \emph{Proceedings of the national academy of sciences}, 116(10): 4156--4165.

\bibitem[{Li et~al.(2020)Li, Rao, Solares, Hassaine, Ramakrishnan, Canoy, Zhu, Rahimi, and Salimi-Khorshidi}]{li2020behrt}
Li, Y.; Rao, S.; Solares, J. R.~A.; Hassaine, A.; Ramakrishnan, R.; Canoy, D.; Zhu, Y.; Rahimi, K.; and Salimi-Khorshidi, G. 2020.
\newblock BEHRT: transformer for electronic health records.
\newblock \emph{Scientific reports}, 10(1): 7155.

\bibitem[{Ma et~al.(2018{\natexlab{a}})Ma, Gao, Suo, You, Zhou, and Zhang}]{ma2018risk}
Ma, F.; Gao, J.; Suo, Q.; You, Q.; Zhou, J.; and Zhang, A. 2018{\natexlab{a}}.
\newblock Risk Prediction on Electronic Health Records with Prior Medical Knowledge.
\newblock In Guo, Y.; and Farooq, F., eds., \emph{Proceedings of the 24th {ACM} {SIGKDD} International Conference on Knowledge Discovery {\&} Data Mining, {KDD} 2018, London, UK, August 19-23, 2018}, 1910--1919.

\bibitem[{Ma et~al.(2018{\natexlab{b}})Ma, You, Xiao, Chitta, Zhou, and Gao}]{ma2018kame}
Ma, F.; You, Q.; Xiao, H.; Chitta, R.; Zhou, J.; and Gao, J. 2018{\natexlab{b}}.
\newblock {KAME:} Knowledge-based Attention Model for Diagnosis Prediction in Healthcare.
\newblock In Cuzzocrea, A.; Allan, J.; Paton, N.~W.; Srivastava, D.; Agrawal, R.; Broder, A.~Z.; Zaki, M.~J.; Candan, K.~S.; Labrinidis, A.; Schuster, A.; and Wang, H., eds., \emph{Proceedings of the 27th {ACM} International Conference on Information and Knowledge Management, {CIKM} 2018, Torino, Italy, October 22-26, 2018}, 743--752.

\bibitem[{Murahari et~al.(2020)Murahari, Batra, Parikh, and Das}]{murahari2020large}
Murahari, V.; Batra, D.; Parikh, D.; and Das, A. 2020.
\newblock Large-scale pretraining for visual dialog: A simple state-of-the-art baseline.
\newblock In \emph{European Conference on Computer Vision}, 336--352. Springer.

\bibitem[{Pfeffer et~al.(2021)Pfeffer, Claggett, Lewis, Granger, K{\o}ber, Maggioni, Mann, McMurray, Rouleau, Solomon et~al.}]{pfeffer2021angiotensin}
Pfeffer, M.~A.; Claggett, B.; Lewis, E.~F.; Granger, C.~B.; K{\o}ber, L.; Maggioni, A.~P.; Mann, D.~L.; McMurray, J.~J.; Rouleau, J.-L.; Solomon, S.~D.; et~al. 2021.
\newblock Angiotensin receptor--neprilysin inhibition in acute myocardial infarction.
\newblock \emph{New England Journal of Medicine}, 385(20): 1845--1855.

\bibitem[{Rasmy et~al.(2021)Rasmy, Xiang, Xie, Tao, and Zhi}]{rasmy2021med}
Rasmy, L.; Xiang, Y.; Xie, Z.; Tao, C.; and Zhi, D. 2021.
\newblock Med-BERT: pretrained contextualized embeddings on large-scale structured electronic health records for disease prediction.
\newblock \emph{NPJ digital medicine}, 4(1): 1--13.

\bibitem[{Rubin(2005)}]{rubin2005causal}
Rubin, D.~B. 2005.
\newblock Causal inference using potential outcomes: Design, modeling, decisions.
\newblock \emph{Journal of the American Statistical Association}, 100(469): 322--331.

\bibitem[{Sandner et~al.(2020)Sandner, Schunkert, Kastrati, Wiedemann, Misfeld, Boening, Tebbe, Nowak, Stritzke, Laufer et~al.}]{sandner2020ticagrelor}
Sandner, S.~E.; Schunkert, H.; Kastrati, A.; Wiedemann, D.; Misfeld, M.; Boening, A.; Tebbe, U.; Nowak, B.; Stritzke, J.; Laufer, G.; et~al. 2020.
\newblock Ticagrelor monotherapy versus aspirin in patients undergoing multiple arterial or single arterial coronary artery bypass grafting: insights from the TiCAB trial.
\newblock \emph{European Journal of Cardio-Thoracic Surgery}, 57(4): 732--739.

\bibitem[{Shalit, Johansson, and Sontag(2017)}]{shalit2017estimating}
Shalit, U.; Johansson, F.~D.; and Sontag, D.~A. 2017.
\newblock Estimating individual treatment effect: generalization bounds and algorithms.
\newblock In Precup, D.; and Teh, Y.~W., eds., \emph{Proceedings of the 34th International Conference on Machine Learning, {ICML} 2017, Sydney, NSW, Australia, 6-11 August 2017}, volume~70 of \emph{Proceedings of Machine Learning Research}, 3076--3085.

\bibitem[{Shi, Blei, and Veitch(2019)}]{shi2019adapting}
Shi, C.; Blei, D.~M.; and Veitch, V. 2019.
\newblock Adapting Neural Networks for the Estimation of Treatment Effects.
\newblock In Wallach, H.~M.; Larochelle, H.; Beygelzimer, A.; d'Alch{\'{e}}{-}Buc, F.; Fox, E.~B.; and Garnett, R., eds., \emph{Advances in Neural Information Processing Systems 32: Annual Conference on Neural Information Processing Systems 2019, NeurIPS 2019, December 8-14, 2019, Vancouver, BC, Canada}, 2503--2513.

\bibitem[{Stolk et~al.(2017)Stolk, de~Vries, Ebbelaar, de~Boer, Schalekamp, Souverein, ten Cate-Hoek, and Burden}]{stolk2017risk}
Stolk, L.~M.; de~Vries, F.; Ebbelaar, C.; de~Boer, A.; Schalekamp, T.; Souverein, P.; ten Cate-Hoek, A.; and Burden, A.~M. 2017.
\newblock Risk of myocardial infarction in patients with atrial fibrillation using vitamin K antagonists, aspirin or direct acting oral anticoagulants.
\newblock \emph{British journal of clinical pharmacology}, 83(8): 1835--1843.

\bibitem[{Vaswani et~al.(2017)Vaswani, Shazeer, Parmar, Uszkoreit, Jones, Gomez, Kaiser, and Polosukhin}]{vaswani2017attention}
Vaswani, A.; Shazeer, N.; Parmar, N.; Uszkoreit, J.; Jones, L.; Gomez, A.~N.; Kaiser, L.; and Polosukhin, I. 2017.
\newblock Attention is All you Need.
\newblock In Guyon, I.; von Luxburg, U.; Bengio, S.; Wallach, H.~M.; Fergus, R.; Vishwanathan, S. V.~N.; and Garnett, R., eds., \emph{Advances in Neural Information Processing Systems 30: Annual Conference on Neural Information Processing Systems 2017, December 4-9, 2017, Long Beach, CA, {USA}}, 5998--6008.

\bibitem[{Xu et~al.(2023)Xu, Chu, Yang, Wang, Zou, Ding, Zhao, Wang, and Xie}]{xu2023seqcare}
Xu, Y.; Chu, X.; Yang, K.; Wang, Z.; Zou, P.; Ding, H.; Zhao, J.; Wang, Y.; and Xie, B. 2023.
\newblock SeqCare: Sequential Training with External Medical Knowledge Graph for Diagnosis Prediction in Healthcare Data.
\newblock In \emph{Proceedings of the ACM Web Conference 2023}, 2819--2830.

\bibitem[{Yang et~al.(2015)Yang, Yih, He, Gao, and Deng}]{yang2014embedding}
Yang, B.; Yih, W.; He, X.; Gao, J.; and Deng, L. 2015.
\newblock Embedding Entities and Relations for Learning and Inference in Knowledge Bases.
\newblock In Bengio, Y.; and LeCun, Y., eds., \emph{3rd International Conference on Learning Representations, {ICLR} 2015, San Diego, CA, USA, May 7-9, 2015, Conference Track Proceedings}.

\bibitem[{Yang et~al.(2023)Yang, Xu, Zou, Ding, Zhao, Wang, and Xie}]{yang2023kerprint}
Yang, K.; Xu, Y.; Zou, P.; Ding, H.; Zhao, J.; Wang, Y.; and Xie, B. 2023.
\newblock KerPrint: Local-Global Knowledge Graph Enhanced Diagnosis Prediction for Retrospective and Prospective Interpretations.
\newblock In \emph{Proceedings of the AAAI Conference on Artificial Intelligence}, 5357--5365.

\bibitem[{Yasunaga, Leskovec, and Liang(2022)}]{yasunaga2022linkbert}
Yasunaga, M.; Leskovec, J.; and Liang, P. 2022.
\newblock {L}ink{BERT}: Pretraining Language Models with Document Links.
\newblock In \emph{Proceedings of the 60th Annual Meeting of the Association for Computational Linguistics (Volume 1: Long Papers)}, 8003--8016.

\bibitem[{Ye et~al.(2021)Ye, Cui, Wang, Luo, Xiao, and Ma}]{ye2021medpath}
Ye, M.; Cui, S.; Wang, Y.; Luo, J.; Xiao, C.; and Ma, F. 2021.
\newblock Medpath: Augmenting health risk prediction via medical knowledge paths.
\newblock In \emph{Proceedings of the Web Conference 2021}, 1397--1409.

\bibitem[{Yusuf et~al.(2004)Yusuf, Hawken, {\^O}unpuu, Dans, Avezum, Lanas, McQueen, Budaj, Pais, Varigos et~al.}]{yusuf2004effect}
Yusuf, S.; Hawken, S.; {\^O}unpuu, S.; Dans, T.; Avezum, A.; Lanas, F.; McQueen, M.; Budaj, A.; Pais, P.; Varigos, J.; et~al. 2004.
\newblock Effect of potentially modifiable risk factors associated with myocardial infarction in 52 countries (the INTERHEART study): case-control study.
\newblock \emph{The lancet}, 364(9438): 937--952.

\bibitem[{Zhang et~al.(2022)Zhang, Zhang, Lipton, Li, and Xing}]{zhang2022can}
Zhang, Y.-F.; Zhang, H.; Lipton, Z.~C.; Li, L.~E.; and Xing, E.~P. 2022.
\newblock Can Transformers be Strong Treatment Effect Estimators?
\newblock \emph{ArXiv preprint}, abs/2202.01336.

\end{thebibliography}

\clearpage

\appendix
\renewcommand{\thefigure}{A\arabic{figure}}
\setcounter{figure}{0}
\renewcommand{\thetable}{A\arabic{table}}
\setcounter{table}{0}
\setcounter{equation}{0}
\setcounter{footnote}{0}
\setlength{\tabcolsep}{6pt}

\section{A \quad Causal Assumption}
Our study is based on three standard assumptions in causal inference. We elaborate each of them as below:
\begin{assumption}[Consistency]\label{as:consistency}
The potential outcome under the treatment $A$ equals the observed outcome if the actual treatment is $A$.
\end{assumption}
\begin{assumption}[Positivity]\label{as:positivity}
Given the observational patient data and treatment-covariate PKG, if $P(A=1|X=\boldsymbol{x}, G=g_a)\neq0$, then 
$0<P(A=a|X=\boldsymbol{x}, G=g_a)<1$, for $a\in\{0, 1\}$ and $\boldsymbol{x}\in\mathcal{X}$.   
\end{assumption}
\begin{assumption}[Strong Ignorability]\label{as:ignorability}
Given the observational patient data and PKGs, the treatment assignment is independent of the potential outcome, i.e.,  $Y(A=a)\independent A | \boldsymbol{X}, g_a, g_y$, for $a\in\{0, 1\}$. 
\end{assumption}
Assumption \ref{as:consistency} is fundamental to the potential outcome framework \citep{rubin2005causal}, which defines factual outcomes, counterfactual outcomes, and treatment effects. The assumption requires that the treatment specified in the study is precise enough that any variation within the treatment specification will not lead to a different outcome. Assumption \ref{as:positivity} suggests that all patients, regardless of their covariates, have the potential to receive the treatment. Without this assumption, it would be impossible to derive counterfactuals for patients who do not have any chance of being in the other treatment group. Assumption \ref{as:ignorability} states that potential outcomes are independent of the treatment assignment, given the set of observed covariates and relevant PKGs.

\section{B \quad Transformer Architecture}
\label{sup:transformer}
The Transformer architecture plays a crucial role in our approach due to its ability to handle sequence data effectively. This is particularly relevant in analyzing patient data where temporal patterns can provide significant insights.
Each single Transformer encoder block consists of a multi-head self-attention layer followed by a fully-connected feed-forward layer \citep{vaswani2017attention}. The multi-head attention is the most crucial part which can be calculated as:
\begin{equation}
    \begin{aligned}
        \text{MultiHead}(\boldsymbol{h})&=\text{Concat}(\text{head}_1,\ldots,\text{head}_h)W^{O},\\
        \text{head}_i&=\text{Attention}(\boldsymbol{h}W^{Q}_i, \boldsymbol{h}W^{K}_i, \boldsymbol{h}W^{V}_i),\\
        \text{Attention}(Q,K,V)&=\text{Softmax}(\frac{QK^{T}}{\sqrt{d}})V,
    \end{aligned}
\end{equation}
where $\boldsymbol{h}\in\mathbb{R}^{d\times d_\text{model}}$ denotes the hidden representations and $d$ is the input sequence length. $W^{Q}_i\in\mathbb{R}^{d_\text{model}\times d}$ , $W^{K}_i\in\mathbb{R}^{d_\text{model}\times d}$ , $W^{V}_i\in\mathbb{R}^{d_\text{model}\times d}$, $W^{O}\in\mathbb{R}^{nd\times d_\text{intermediate}}$ are learnable parameter matrices. $d=d_\text{model}/n$ and $n$ is the number of attention heads.

\section{C \quad Additional Experimental Setup}
\label{sup:exp_setup}
\begin{figure}[!ht]
\centering
\includegraphics[width=0.9\linewidth]{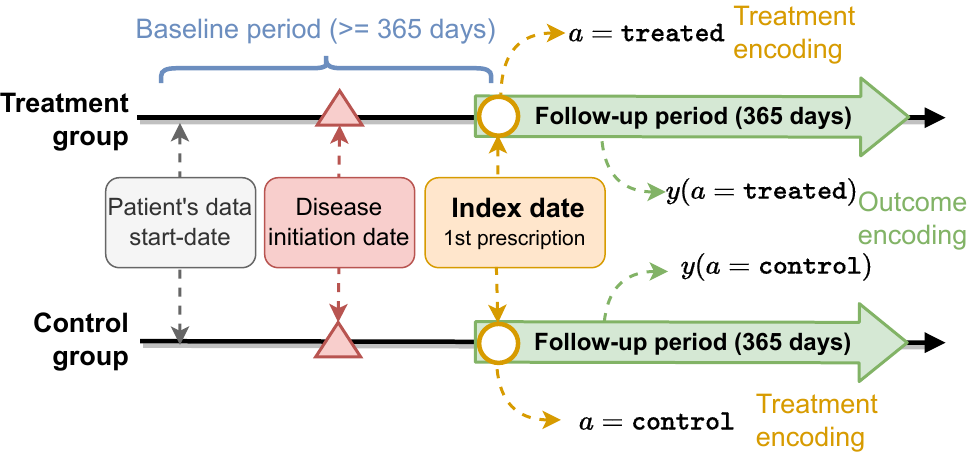}
\caption[]{
Illustration of the downstream data construction for treatment and control patient groups. The index date refers to the first prescription of either the treated or control medication, which should not precede the disease (CAD) initiation date. The baseline period before the index date is set to be no less than one year and the follow-up period after the index date is also set to one year.
}
\label{fig:study_design}
\end{figure}

\subsection{Pre-training Data} 
We obtain patient data from the MarketScan Commercial Database\footnote{https://www.merative.com/real-world-evidence}, which consists of medical and drug data from employers and health plans for over 215 million individuals. In this paper, we extract millions of patients who have received a diagnosis of coronary artery disease (CAD) as our pre-training patient data. The medical definition of CAD is provided in Table \ref{tab:cad_definition}.

\begin{table*}[!t]
\centering
\begin{tabular}{ll}
\toprule
Reference (PMID) & \begin{tabular}[c]{@{}l@{}}16159046, 26524702, 28008010 \end{tabular} \\ \midrule
Criteria &
  \begin{tabular}[c]{@{}l@{}}A history of coronary revascularization in the EHR\\ Or, history of acute coronary syndrome, ischemic heart disease, or exertional angina\end{tabular} \\ \midrule
Diagnostic codes &
  \begin{tabular}[c]{@{}l@{}}ICD-9 codes:\\ 411* to 414*\\ ICD-10 codes:\\ The best approximation are the following codes:\\ I20* Angina pectoris\\ I22* Subsequent ST elevation (STEMI) and non-ST elevation (NSTEMI) myocardial\\ infarction\\ I23* Certain current complications following ST elevation (STEMI) and non-ST\\ elevation (NSTEMI) myocardial infarction (within the 28 day period)\\ I24* Other acute ischemic heart diseases\\ I25* Chronic ischemic heart disease\end{tabular} \\ \bottomrule
\end{tabular}
\caption{The definition of coronary artery disease (CAD) from observational health data.}
\label{tab:cad_definition}
\end{table*}

\begin{table*}[!ht]
\centering
\begin{tabular}{ll}
\toprule
Reference (PMID) & \begin{tabular}[c]{@{}l@{}}29202795 \end{tabular} \\ \midrule
Diagnostic codes &
  \begin{tabular}[c]{@{}l@{}}ICD-9 codes:\\ V12.54,\\ 438.0–438.9\\ 410* \\ ICD 10 codes:\\ Z86.73\\ I60-I69\\ subarachnoid hemorrhage (I60);\\ intracerebral hemorrhage (I61);\\ cerebral infarction (I63);\\ and other transient cerebral ischemic attacks and related syndromes and\\ transient cerebral ischemic attack (unspecified) (G458 and G459) \\ I21* Acute myocardial infarction\end{tabular} \\ \bottomrule
\end{tabular}
\caption{The definition of stroke and myocardial infarction from observational health data}
\label{tab:stroke_definition}
\end{table*}

\subsection{Downstream Tasks}
The goal of the downstream tasks is to evaluate the effects of
two treatments in reducing the risk of stroke and myocardial infarction given the patient's observational data. In Fig. \ref{fig:study_design}, we illustrate the process of deriving each downstream dataset for a given treatment-control medication pair. Note that patient covariates are obtained from the baseline period before the first prescription of the treatment or control medication. Outcomes are then collected from the follow-up period to compare the effectiveness of the treatments.

\begin{figure}[!ht]
\centering
\includegraphics[width=0.9\linewidth]{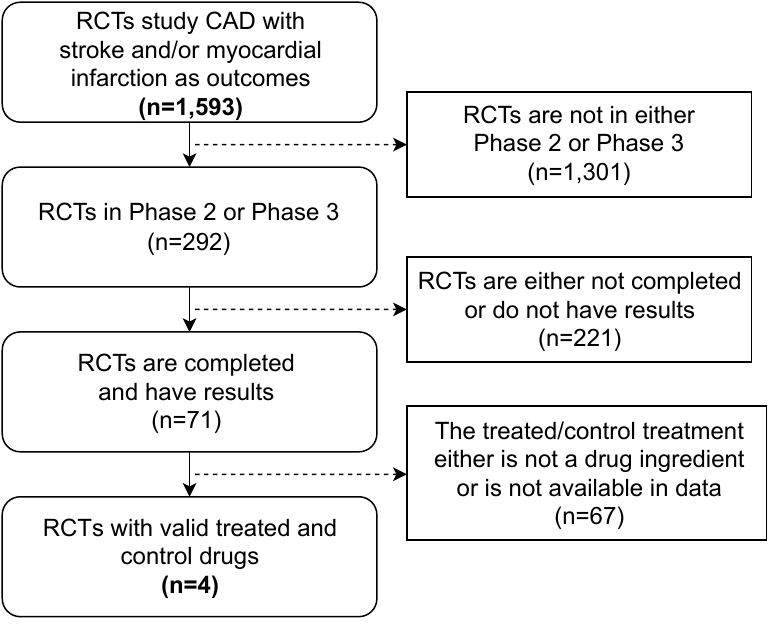}
\caption[]{The data flow for RCT extraction. The downstream tasks are constructed based on the extracted RCTs.}
\label{fig:downstream_data}
\end{figure}

As the ground truth treatment effects are not available in observational data and RCTs are
the gold standard for TEE, we create downstream datasets by examining CAD-related RCTs obtained from \url{https://clinicaltrials.gov/}. As shown in Fig. \ref{fig:downstream_data}, we start with 1,593 RCTs that study CAD with stroke and/or myocardial infarction as the outcomes, and end up with 4 RCTs that satisfy all the criteria. Finally, we construct 4 downstream datasets corresponding to each RCT study. The statistics of 4 datasets are provided in Table \ref{tab:downstream_stat}.

\begin{table*}[!ht]
\centering
\adjustbox{max width=\textwidth}{
\begin{tabular}{lcccc}
\toprule
 Target v.s. Compared & Rivaroxaban v.s. Aspirin & Valsartan v.s. Ramipril & Ticagrelor v.s. Aspirin & Apixaban v.s. Warfarin \\ \midrule
\# of patients (Target; Compared) & 26340 (9569; 16771)      & 12850 (7306; 5544)      & 29248 (12477; 16771)    & 18187 (6701; 11486)    \\
Female (\%)                      & 30.4  & 32.4  & 27.1  & 31.8  \\
Age (group) on index date        & 55-64 & 55-64 & 55-64 & 55-64 \\
Patients with stroke (\%)        & 13.7  & 11.9  & 18.9  & 16.7  \\
Average \# of visits per patient & 83.4  & 74.0  & 70.7  & 97.1  \\
Average \# of codes per patient  & 206.4 & 177.2 & 173.1 & 244.5 \\ \bottomrule
\end{tabular}}
\caption{The statistics of 4 downstream datasets.}
\label{tab:downstream_stat}
\end{table*}

\subsection{Evaluation Metrics}
We use influence function-based precision of estimating heterogeneous effects (IF-PEHE) \citep{alaa2019validating} to evaluate the accuracy in effect estimation. We elaborate the computation procedure of IF-PEHE as follows:
\begin{itemize}[leftmargin=*]
    \item Step 1: Train two XGBoost \citep{chen2016xgboost} classifiers for potential outcome prediction denoted by $\mu_0$ and $\mu_1$, where $\mu_{a}=P(y(a)=1|X=x)$ using the training set $\mathcal{Z}_\text{train}$. Then calculate the plug-in estimation $\widetilde{T}=\mu_1-\mu_0$
    Train an XGBoost \citep{chen2016xgboost} classifier propensity score function (i.e., the probability of receiving treatment) $\widetilde{\pi}=P(a=1|X=x)$.
    \item Step 2: Given the estimated treatment effect $\hat{T}(x_i)$ on the test set $\mathcal{Z}_{\text{test}}$, calculate the IF-PEHE with the influence function $\hat{l}$ as:
    \begin{gather*}
        \text{IF-PEHE}=\sum_{x_i\in\mathcal{Z}_{\text{test}}}[(\hat{T}(x_i)-\widetilde{T}(x_i))^2+\hat{l}(x_i)], \\
        \hat{l}(x)=(1-B)\widetilde{T}^{2}(x)+(\widetilde{T}(x)-\hat{T}(x))-W(\widetilde{T}(x)-\hat{T}(x))^2\\ + \hat{T}^{2}(x),
    \end{gather*}
where $W=(a-\widetilde{\pi}(x))$, $B=2a(a-\widetilde{\pi}(x))C^{-1}$, $C=\widetilde{\pi}(x)(1-\widetilde{\pi}(x))$.
\end{itemize}

\subsection{Implementation Details}

In our model, we utilize the BERT-base architecture \citep{devlin2018bert} for the patient sequence encoder. This architecture consists of 12 Transformer layers, each with 12 attention heads, and uses a hidden size of 768 and an intermediate size of 3072. Patient Knowledge Graphs (PKGs) are retrieved using a 2-hop bridge node method. The node embeddings in these PKGs are initialized with pooled embeddings from BioLinkBERT \cite{yasunaga2022linkbert}, a specialized model trained on biological data. The Graph Neural Network (GNN) encoder, which processes the PKGs, has a hidden size of 200. The number of deep bi-level attention synergy layers is 5. For the KG link prediction, we use the DistMult scoring function \cite{yang2014embedding}, together with 64 corrupted triplets, to optimize the learning of KG relationships. The pre-training process is conducted on 4 NVIDIA GeForce RTX 2080 Ti 16GB GPUs. We use the Adam optimizer given its ability to adapt the learning rate for each parameter, which can lead to more efficient and robust model training. The downstream data is split into training, validation, and test sets in a 90\%, 5\%, 5\% ratio, respectively. This split ensures that our model is evaluated on unseen data, thus providing a more reliable measure of its performance. All results reported in this paper are based on the test sets. More detailed hyperparameters setup is shown in Table \ref{tab:pretrain_parameter} for pre-training, and Table \ref{tab:finetune_parameter} for fine-tuning. 

\begin{table}[!ht]
\centering
\begin{tabular}{lc}
\toprule
Parameters            & \model \\ \midrule
Maximum Steps         & 200K                  \\
Initial Learning Rate & 1e-4                  \\
Batch Size            & 28                    \\
Warm-Up Steps         & 20K                   \\
Sequence Length       & 256                 \\
Number of nodes       & 200                 \\
Dropout               & 0.1                   \\ \bottomrule
\end{tabular}
\caption{Hyperparameters used in pre-training.}
\label{tab:pretrain_parameter}
\end{table}

\begin{table}[!ht]
\centering
\adjustbox{max width=\linewidth}{
\begin{tabular}{lcc}
\toprule
Parameters            & Search Space          & Optimal Value \\ \midrule
Maximum Epochs        & \{2,4,8,16\}         & 2             \\
Initial Learning Rate & \{1e-5, 3e-5, 5e-5\}  & 5e-5          \\
Batch Size            & \{8, 16, 32\}        & 32            \\
Sequence Length       & 256                   & 256           \\
Fixed Window Length       & 30                   & 30           \\
Baseline Window       & \{90, 180, 360, 720\} & 360           \\
Dropout               & 0.1                   & 0.1           \\ 
$\beta$ & \{0, 0.5, 1, 1.5, 2\} & 1\\
\bottomrule
\end{tabular}}
\caption{The search space of hyperparameters and the optimal parameters utilized during the fine-tuning process.}
\label{tab:finetune_parameter}
\end{table}

\section{D \quad Additional Results}

\subsection{Qualitative Analysis}
A thorough comparison of \model with state-of-the-art methods in terms of validation with randomized controlled trials (RCTs) conclusions is provided in Table \ref{tab:rct_all}. The results show that \model is more effective in accurately estimating treatment effects and generating consistent conclusions with RCTs than the baseline methods. 

\subsection{Ablation Study: Downstream Data Size}
\begin{figure}[!ht]
\centering
\includegraphics[width=0.99\linewidth]{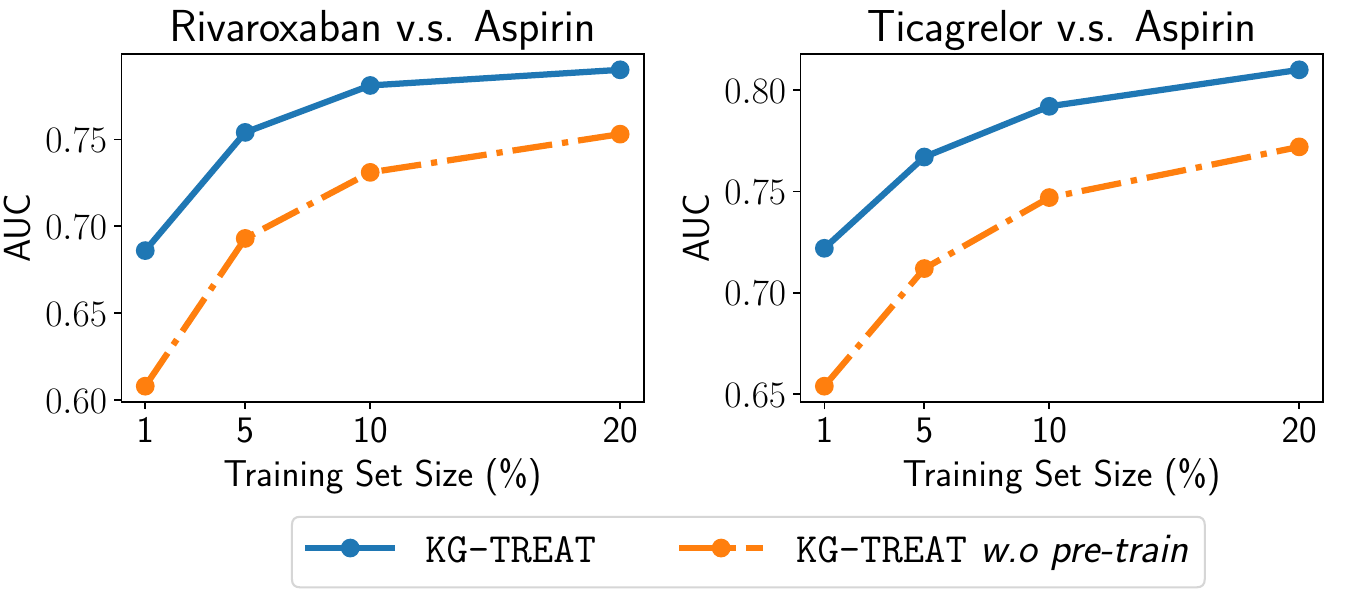}
\caption{Performance of low-resource fine-tuning data on Rivaroxaban v.s. Aspirin and Ticagrelor v.s. Aspirin datasets with different fractions (\%) of the training set used (x-axes).}
\label{fig:ablation_few_shot}
\end{figure}

In Fig. \ref{fig:ablation_few_shot}, we show the model performance on low-resource fine-tuning data with training set fractions of $\{1\%, 5\%, 10\%, 20\%\}$. Compared to the base model, \model \textit{w/o pre-train}, that directly trained on the downstream data, \model demonstrates larger performance gains especially given only limited fine-tuning data samples (e.g., the model attains 7-8\% AUC improvements using only 1\% training data in 2 downstream tasks). The results suggest the generalizability of pre-training and fine-tuning framework in limited data for TEE.

\begin{table*}[!ht]
\centering
\adjustbox{max width=\textwidth}{
\begin{tabular}{l|cccccc}\toprule
\multirow{2}{*}{Method} & \multicolumn{3}{c}{Rivaroxaban v.s. Aspirin} &
  \multicolumn{3}{c}{Valsartan v.s. Ramipril} \\     & Estimated Effect (CI)    & P value   & Match RCT Conclusion? & Estimated Effect (CI)   & P value   & Match RCT Conclusion? \\ \midrule
TARNet 
& [0.066, 0.095]           & 5.678e-10 & No                    & [-0.037, -0.003]        & 0.026     & No                    \\
DragonNet 
& [0.18, 0.236]            & 5.979e-12 & No                    & [0.03, 0.07]            & 4.681e-5 & No                    \\
DR-CFR 
& [0.13, 0.183]            & 2.783e-10 & No                    & [0.002, 0.04]           & 0.033     & No                    \\
TNet 
& [0.041, 0.07]            & 2.509e-7 & No                    & [-0.038, -0.001]        & 0.039     & No                    \\
SNet 
& [-0.002, 0.008]          & 0.231     & Yes                   & [-0.051, -0.026]        & 3.168e-6 & No                    \\
FlexTENet 
& [0.064, 0.108]           & 1.529e-7 & No                    & [-0.079, -0.035]        & 3.184e-5 & No                    \\
TransTEE 
& [-0.013, -0.002]         & 0.018     & No                    & [-0.019, 0.034]         & 0.420     & Yes                   \\ \midrule
\model      & [-0.010, 0.009]          & 0.952     & Yes                   & [-0.015, 0.007]         & 0.564     & Yes                   \\ \bottomrule \toprule
\multirow{2}{*}{Method} &            \multicolumn{3}{c}{Ticagrelor v.s. Aspirin} &
  \multicolumn{3}{c}{Apixaban v.s. Warfarin}        \\
& Estimated Effect (CI)    & P value   & Match RCT Conclusion? & Estimated Effect (CI)   & P value   & Match RCT Conclusion? \\ \midrule
TARNet 
& [0.064, 0.101]           & 2.861e-8 & No                    & [-0.006, 0.028]         & 0.207     & No                    \\
DragonNet 
& [-0.013, 0.01]           & 0.821     & Yes                   & [0.018, 0.056]          & 6.284e-4 & No                    \\
DR-CFR 
& [-0.068, -0.029]         & 4.915e-5 & No                    & [-0.026, -0.002]        & 0.047     & Yes                   \\
TNet 
& [0.046, 0.069]           & 6.474e-9 & No                    & [0.009, 0.023]          & 2.329e-4 & No                    \\
SNet 
& [0.005, 0.016]           & 4.398e-4 & No                    & [-0.046, -0.017]        & 2.112e-4 & Yes                   \\
FlexTENet 
& [0.045, 0.068]           & 5.243e-9 & No                    & [0.012, 0.042]          & 0.001     & No                    \\
TransTEE 
& [-0.014, -0.009]         & 0.0216    & No                    & [-0.027, -0.002]        & 0.027     & Yes                   \\ \midrule
\model      & [-0.006, 0.021]           & 0.436 & Yes                    & [-0.006, -0.001]        & 0.001     & Yes   \\ \bottomrule               
\end{tabular}}
\caption{Comparison of the estimated treatment effects of all methods with corresponding RCT conclusions.}
\label{tab:rct_all}
\end{table*}

\end{document}